\pdfoutput=1

\documentclass[11pt]{article}

\usepackage[preprint]{acl}

\usepackage{times}
\usepackage{latexsym}

\usepackage{booktabs} 
\usepackage{booktabs}   
\usepackage{multirow}   
\usepackage{makecell}   
\usepackage{caption}    
\usepackage[most]{tcolorbox}

\usepackage{subcaption}  
\usepackage{amsmath,amssymb}
\newtcolorbox{promptbox}[2][]{breakable,title=#2,#1}

\usepackage{graphicx}
 \usepackage{booktabs}       
 \usepackage{float}          
 \usepackage{tabularx}       
 \usepackage{amsmath}        
 \usepackage{multirow}       
\usepackage[T1]{fontenc}

\usepackage[utf8]{inputenc}

\usepackage{microtype}

\usepackage{inconsolata}

\usepackage{graphicx}

\title{RL Teaches VLMs Where to Attend but Not How to Perceive}

\author{
  Ziqian Zhang\textsuperscript{1},
  Junjie Liao\textsuperscript{2},
  Zhining Zhang\textsuperscript{3},
  Wentao Zhu\textsuperscript{4}\thanks{Correspondence to: Wentao Zhu (\texttt{wtzhu@pku.edu.cn}).} \\
  \textsuperscript{1}National University of Singapore \\
  \textsuperscript{2}Beijing Normal University \\
  \textsuperscript{3}Peking University \\
  \textsuperscript{4}Eastern Institute of Technology, Ningbo \\
  \texttt{e1546905@u.nus.edu},
  \texttt{wtzhu@pku.edu.cn}
}

\begin{document}
\maketitle
\begin{abstract}
 Although reinforcement learning (RL) effectively trains Vision-Language Model (VLM) agents, the mechanisms behind these capability gains are poorly understood. Across four diverse agent environments, we investigate whether the task-success gains from multi-turn RL are accompanied by improved grounded visual question answering about the environment. To assess this, we design environment-specific grounded VQA tasks that reveal improvements in grounded visual understanding. Representational probing shows that vision-encoder probe accuracy remains relatively stable, while LLM-layer probe accuracy improves. Attention analysis further reveals that LLM layers pay more attention to visual tokens during RL training. To test whether increased attention to visual tokens contributes to VQA gains, we introduce a lightweight plug-and-play intervention using a fixed attention bias. Without updating model parameters or computing gradients, this intervention partially restores VQA gains on the untrained model. These results reveal better utilization of visual information as a key mechanism behind the gains from RL.
\end{abstract}


\section{Introduction}
\label{sec:intro}

\begin{figure*}[t]
  \centering
  \includegraphics[width=\textwidth]{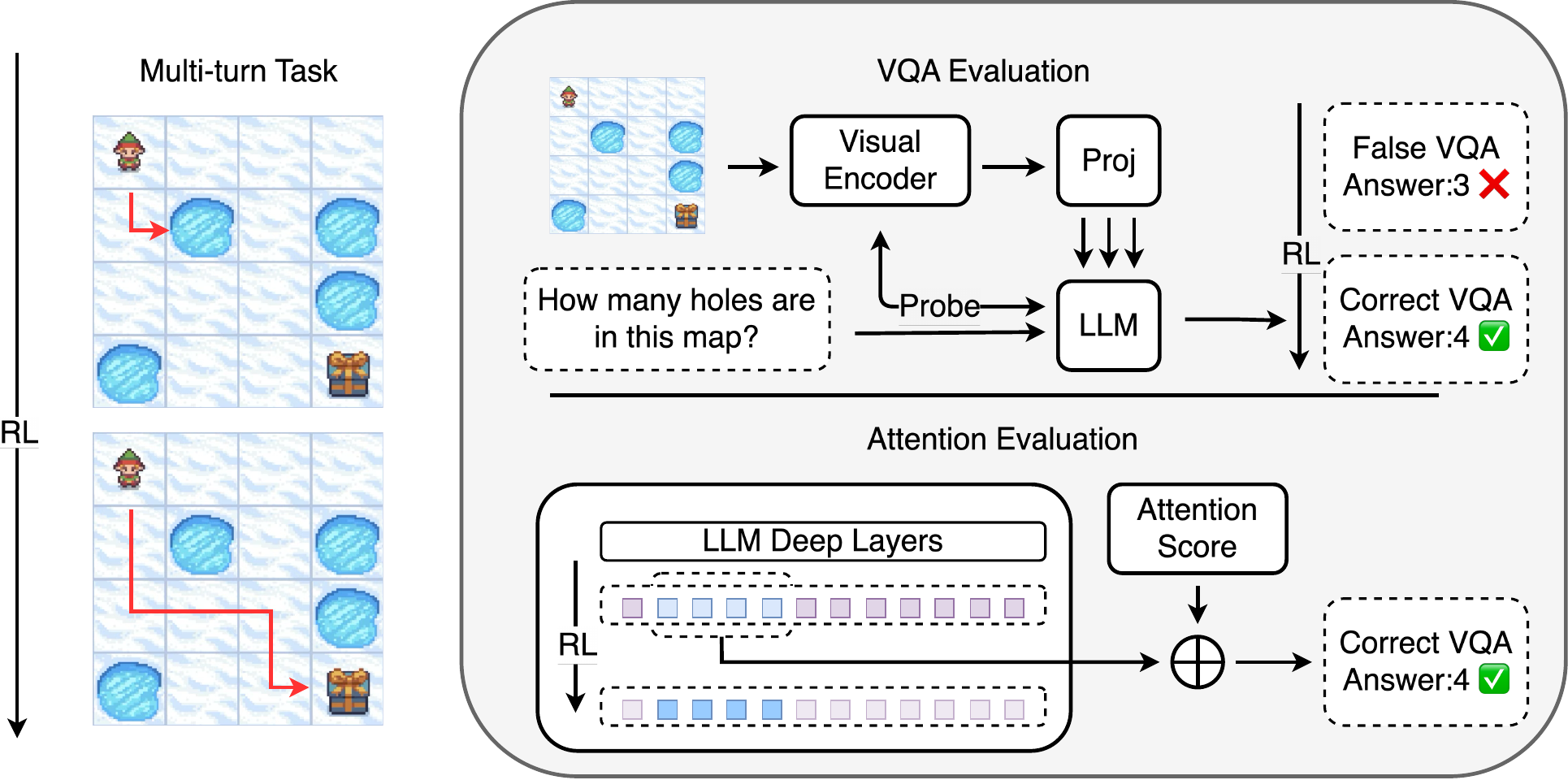}
  \caption{\textbf{Overview of our evaluation framework and attention bias intervention.}
  Left: RL progressively improves multi-turn task performance in FrozenLake.
  Top right: VQA evaluation with layer-wise probing reveals that RL enhances visual utilization in deep LLM layers.
  Bottom right: RL training deepens the attention weights on visual tokens (darker colors) in deep LLM layers;
  injecting an additive attention bias into the base model partially replicates this shift
  and improves VQA accuracy.}
  \label{fig:method}
\end{figure*}
Vision-Language Models (VLMs) have emerged as foundational infrastructure
for cross-modal reasoning, owing to pre-training on large-scale
image-text paired data~\cite{li2022blip,radford2021learning}.
As applications expand from static VQA and image captioning toward
Embodied AI and multi-turn decision-making
agents~\cite{savva2019habitat,kolve2017ai2,xi2025agentgym,li2024embodied,ALFWorld20},
agents must actively perceive environmental changes, comprehend
multi-step instructions, and perform sequential planning to accomplish
long-horizon goals~\cite{feng2025group,wang2025vagen}.
This shift exposes the limitations of general-purpose pre-trained VLMs,
whose static image-text training does not fully cover dynamic causality
and sequential decision-making.
Visual instruction tuning datasets similarly offer limited coverage
of these scenarios~\cite{zhai2024fine}.
To bridge this gap, Reinforcement Learning has been widely used for
environment-specific VLM adaptation, alongside optimization methods
such as PPO, GRPO, and
DPO~\cite{schulman2017proximal,shao2024deepseekmath,guo2025deepseek,dpo}.
By converting environmental feedback into optimization signals,
RL reshapes the VLM output distribution to align with target-environment
dynamics, improving task success rates and policy
robustness~\cite{wang2023voyager,li2025search,hong2024cogagent}.

A fundamental question nevertheless remains:
does this macroscopic improvement truly reflect a substantive evolution
in the model's low-level grounded visual reasoning,
or has the agent merely learned better action trajectories?
Although some studies note that RL generalizes better than SFT and enhances visual rewards~\cite{chu2025sft,jin2025rl,chen2025g1bootstrappingperceptionreasoning,liu2025visual},
a direct controlled experiment confirming visual understanding improvement has been absent.
Even if confirmed, the underlying mechanism remains a largely unexplored black box:
does the performance leap originate from enhanced low-level feature extraction
in the Vision Encoder, or from improved attention allocation in the LLM backbone?
Answering this question can clarify how VLM agents use visual information and guide the development of RL methods that improve visual grounding.

To address these questions, we present a comprehensive empirical study of VLMs under multi-turn RL fine-tuning
across four diverse agent environments with environment-specific grounded VQA experiments.
On these grounded VQA tasks, RL-finetuned agents achieve VQA accuracy gains of up to approximately 11 percentage points over the base model.
To disentangle the sources of this improvement,
we apply representational probing at multiple checkpoints to assess information in the output features of the vision encoder and LLM layers. 


The results reveal a counter-intuitive finding:
vision-encoder probe accuracy remains relatively stable throughout RL optimization,
while probe accuracy of LLM intermediate and final layers shows a consistent upward trend.
This points to RL reshaping the LLM backbone's visual information utilization rather than visual perception.
Further investigation traces this improved utilization to greater attention allocation toward visual tokens,
supported by attention intervention experiments:
an additive bias injected at visual token positions in selected Transformer layers
partially replicates the VQA gains achieved by RL training without updating any model parameters,
  supporting this attention shift as a contributing mechanism for visual grounding.
Our intervention diagnoses visual information utilization on static grounded VQA.
Our main contributions are summarized as follows:

\textbf{Revealing capability improvements under multi-turn RL.}
    We confirm that multi-turn RL improves not only task success rates
    but also grounded VQA performance on environment-related visual tasks.
    
  \textbf{Analyzing visual information utilization.}
    Through representational probing and RAPT-based attention analysis, we find evidence that RL improves visual grounding partly through better utilization of visual information in the LLM backbone.

 \textbf{Interventional evidence for the attention shift hypothesis.}
    We introduce an additive attention intervention at inference time that partially reproduces RL-induced grounded VQA gains on the base model,
    supporting the hypothesis that the LLM's attention utilization strategy
    contributes to the grounded VQA improvements observed in our settings.





\section{Related Works}

\subsection{Reinforcement Learning for Vision-Language Models} 
Reinforcement Learning (RL) has emerged as a key paradigm for adapting Vision-Language Models (VLMs) to complex, dynamic environments, particularly in multi-round decision-making tasks~\cite{111}.  Traditional VLM pre-training, often based on static image-text pairs, struggles to establish the understanding of dynamic causal relationships and sequential decision-making capabilities required for such scenarios \cite{ref1}. Prompting-based approaches \cite{wei2022chain,yao2022react,shinn2023reflexion,yao2023tree} partially address this but are bounded by frozen model weights. RL-based methods, such as Proximal Policy Optimization (PPO) \cite{schulman2017proximal}, Group Relative Policy Optimization (GRPO) \cite{shao2024deepseekmath}, and Direct Preference Optimization (DPO) \cite{dpo}, effectively bridge this gap by converting environmental feedback into optimization signals. Recent research has begun to investigate at a mechanistic level how RL enhances VLM capabilities. For example, Song et al.~\cite{ref7} demonstrated that reinforcement learning (specifically DPO), compared to Supervised Fine-Tuning (SFT), can produce stronger and more localized visual representations. 

\subsection{Mechanistic Interpretability of Vision-Language Models} 

Mechanistic Interpretability aims to reverse-engineer deep neural networks to understand their internal workings, which is crucial for enhancing the robustness, generalization, and explainability of VLMs \cite{ref8,zhu2024language}. Early research into the mechanistic interpretability of VLMs \cite{ref66} primarily focused on dissecting attention mechanisms and identifying key components responsible for cross-modal reasoning. For instance, Golovanevsky et al.~\cite{ref9} proposed a framework for causal analysis of VLM behavior through noise-free text-image corruption and evaluation. A significant finding in VLM interpretability research is the ``seeing but not believing'' phenomenon, introduced by Liu et al.~\cite{ref10}. This study introduced Relative Attention per Token (RAPT) as a metric to quantify the attention received by each token type. They demonstrated that although VLMs may internally perceive relevant visual cues, they might fail to fully utilize this information during the decision-making process.

\section{Evaluation Setup}
\label{sec:setup}

We use Qwen2.5-VL-3B~\cite{Qwen2.5-VL} as the primary backbone for our RL and SFT analyses.
The vision encoder partitions input images into patches, extracts sequential visual features,
and passes them through a projector to the LLM backbone for cross-modal reasoning,
generating action decisions or text responses autoregressively.

\subsection{Agent Environments and Tasks}
\label{subsec:environments}

We evaluate on four varied environments covering 2D grid-worlds to a 3D egocentric task, demanding strict spatial perception and planning logic. Specific details are deferred to the Appendix.

\begin{figure*}[t]
  \centering
  \includegraphics[width=\linewidth]{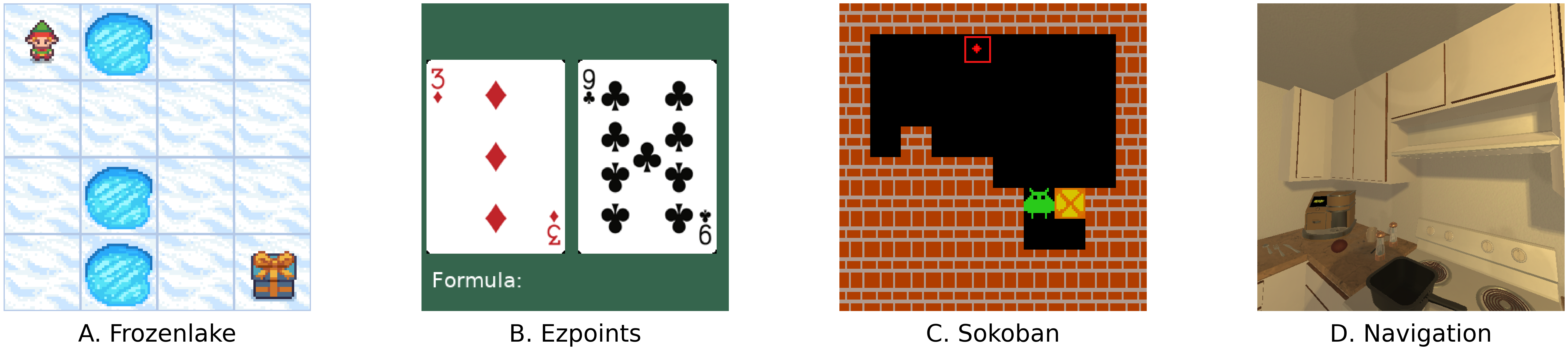}
  \caption{Representative visual observations from the four environments.
  \textbf{A.} FrozenLake.
  \textbf{B.} EzPoints.
  \textbf{C.} Sokoban.
  \textbf{D.} Navigation.}
  \label{fig:env_examples}
\end{figure*}

\paragraph{FrozenLake.} A discrete grid-world task requiring top-down visual navigation to a goal while avoiding visually similar hazards, demanding precise image-to-coordinate mapping.

\paragraph{EzPoints.} A visual arithmetic task where the agent must construct an equation evaluating to 12 from a sequence of observed number cards and operators.

\paragraph{Sokoban.} A classic puzzle introducing strict physical irreversibility, where the agent pushes a box to a target tile, necessitating complex spatial multi-step forward simulation.

\paragraph{Navigation.} A multi-turn 3D ObjectNav task based on the AI2-THOR simulator~\cite{kolve2017ai2}, requiring the agent to search for target objects using egocentric first-person visual inputs.

\subsection{Training Infrastructure and Paradigm Settings}
\label{subsec:training}

Both SFT and RL experiments are initialized from the same Qwen2.5-VL-3B weights and are conducted on an Ubuntu server with A800 GPUs.
The interaction and training pipeline is built on the vagen~\cite{wang2025vagen}
and verl-agent~\cite{feng2025group} frameworks.

\paragraph{Reinforcement Learning Fine-tuning.}
We employ GRPO for multi-turn online interaction.
At each training step, 16 environment instances are sampled in parallel,
generating 8 trajectories each for relative advantage computation.
Training runs for 150 steps.
Reward functions are sparse, with moderate penalties for constraint violations.
Details are in the Appendix.

\paragraph{Supervised Fine-tuning.}
To ensure a data-matched comparison, we fix 2,400 expert trajectories per environment
and train for 4 epochs with full-parameter fine-tuning.
This ensures the gradient update volume of the SFT model is comparable
to data generated during RL exploration,
excluding performance differences due to training data volume alone.
Configurations are provided in the Appendix.

\subsection{VQA, Probing and Intervention Setup}
\label{subsec:probing}

We construct a white-box diagnostic paradigm comprising static evaluation,
representational probing, and interventional validation.

\paragraph{Visual Question Answering Evaluation.}
We convert key information from intermediate environment states into static grounded VQA tasks
that measure changes in low-level visual perception during RL training.
The questions cover hole counting in FrozenLake, card-value recognition in EzPoints, box counting in Sokoban, and object recognition in Navigation.
For each environment, we collect an equal number of images per category,
apply varying visual transformations to ensure diversity,
and split the data into training and test sets.

\paragraph{Layer-level Representational Probing.}
To track representational evolution throughout RL training,
we extract hidden states at three key nodes from a frozen VLM at 11 checkpoints
(checkpoint~0 is the unmodified base model; checkpoints~1--10 are sampled
at regular intervals during the 150-step RL run):
the vision encoder output, an intermediate LLM layer, and the final LLM layer.
A probe classifier is trained on the VQA training set for each node.
Following~\cite{vlmlens}, each probe is a two-layer MLP with ReLU activation
and hidden dimension $h = 512$.
Test-set probe accuracy measures how well task-relevant visual information
can be decoded from each representation.

\paragraph{Attention Tracking and Interventional Validation.}
We extract attention matrices across all LLM layers on the VQA test set
and compute the Relative Attention per Token (RAPT) metric stratified by layer depth.
RAPT is the ratio of average attention received by tokens of a given modality
to the global per-token average; its change over RL training reveals
whether RL induces systematic attention redistribution.
To validate this, we design a plugin that injects layer-wise additive attention biases
at designated layers to enhance visual token attention.
The plugin introduces one fixed constant scalar bias per intervened layer
(by default, the latter half of the LLM layers), without any parameter updates or gradient computations.
We select the specific bias values via grid search, as detailed in Appendix~\ref{sec:appendix_hyper_search}.
This paradigm tests whether targeted visual attention enhancement serves as a contributing mechanism
that contributes to the performance gains observed under RL.


\section{Empirical Study: What Does RL Bring to VLM Agents?}
\label{sec:empirical}

In this section, we present our empirical findings to answer the fundamental question raised in Section~\ref{sec:intro}: Does multi-turn reinforcement learning (RL) fine-tuning merely optimize the agent's action trajectories, or does it genuinely enhance the fundamental grounded visual reasoning capabilities of VLMs?

\begin{figure*}[h]
  \centering
  \includegraphics[width=\linewidth]{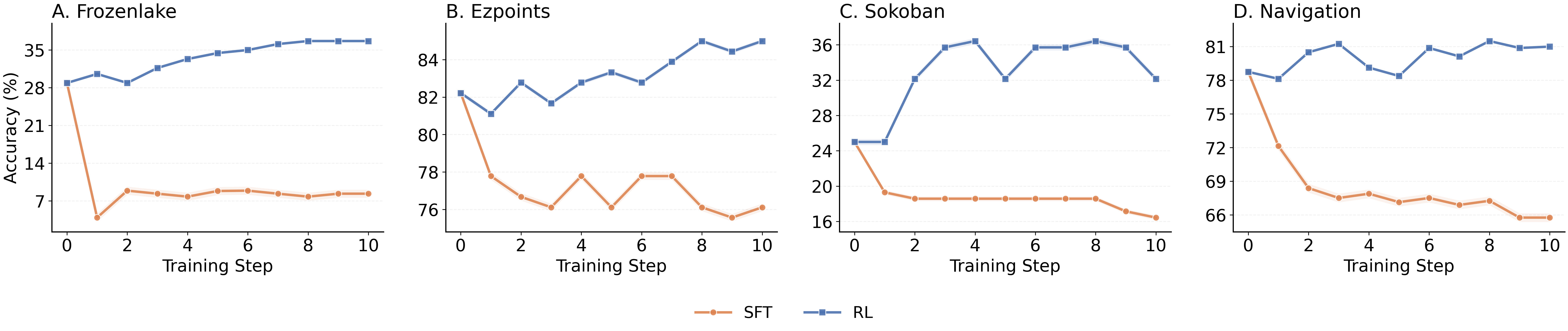}
  \caption{\textbf{Comparison of VQA accuracy for the RL agent and SFT fine-tuning in FrozenLake, EzPoints, Sokoban, and Navigation environments.} Compared to the SFT model (red), the VQA accuracy of the RL agent (blue) achieves greater improvement as the number of training steps increases across all four environments.}
  \label{fig:sft}
\end{figure*}

\subsection{Task Performance}

We first examine the macroscopic impact of RL fine-tuning (optimized via GRPO) on the agent's task completion abilities across four environments: FrozenLake, EzPoints, Sokoban, and Navigation, compared with the zero-shot base model.

As the number of RL training steps increases, the task success rate exhibits a steady and significant upward trend. Specifically, the task success rate increases from 1.56\% to 81.25\% in the EzPoints environment, from 15.62\% to 78.13\% in FrozenLake, and from 13.47\% to 80.35\% in Sokoban, with similarly substantial gains observed in Navigation. These results confirm the effectiveness of the RL paradigm in optimizing multi-turn decision-making.

\subsection{Grounded VQA Performance}

To disentangle the agent's visual understanding capabilities from its policy execution, we evaluate intermediate checkpoints during the RL training process on static Grounded Visual Question Answering (VQA) benchmarks constructed specifically for each environment. These VQA tasks assess the ability to parse spatial relationships, object states, and numerical attributes, independent of the sequential decision-making process.

Figure~\ref{fig:sft} illustrates the evolution of VQA accuracy during the RL optimization process. 

Qwen3-VL-4B provides additional validation in FrozenLake: VQA accuracy rises
from 42.22\% to 51.89\%, while vision-encoder probe accuracy remains stable
and LLM-layer probes improve (Appendix~\ref{sec:appendix_qwen3_rl}).
Cross-model plugin results are reported in Appendix~\ref{sec:appendix_arch}.

    \textbf{Correlation between RL training and VQA capability enhancement.} In FrozenLake, EzPoints, Sokoban, and Navigation, we observe a clear positive correlation trend between grounded VQA accuracy and RL training steps. This indicates that the optimization signals derived from sparse environmental rewards may have successfully backpropagated, contributing to the enhancement of the model's understanding of static images.
    
     \textbf{RL fine-tuning demonstrates potential exceeding SFT in visual grounding.} Although the SFT model is exposed to the exact same visual states during behavioral cloning, SFT does not improve VQA accuracy over the base model in these evaluations, whereas RL improves it. This suggests that exploring the environment and learning from feedback guides the model to establish a more robust visual grounding mechanism than passive supervised learning.
    
    \textbf{The enhancement effect exhibits a consistent trend across environments.} Despite substantial differences in visual complexity and action spaces among the four environments (e.g., discrete spatial navigation in FrozenLake vs. continuous numerical reasoning in EzPoints, or 3D egocentric scenes in Navigation), the improvement patterns in VQA demonstrate strong consistency.

Our empirical results suggest that multi-turn RL fine-tuning does not merely teach VLM agents \emph{how to act}, but likely also enhances their ability to \emph{see} and understand the environment. The performance gap between RL and SFT on static VQA tasks indicates that RL may induce a shift in how the model processes visual information. This macroscopic leap in capability raises a deeper mechanistic question: \emph{Where exactly does this visual enhancement originate within the massive VLM architecture?} Did the sparse RL rewards successfully update the low-level feature extraction capabilities of the Vision Encoder, or did they merely optimize how the LLM backbone routes and attends to these visual features? To answer this question, we conduct a mechanistic analysis, which will be detailed in the next section.

\section{Mechanistic Analysis}
\label{sec:mechanism}

\begin{figure*}[t]
  \centering
  \includegraphics[width=\linewidth]{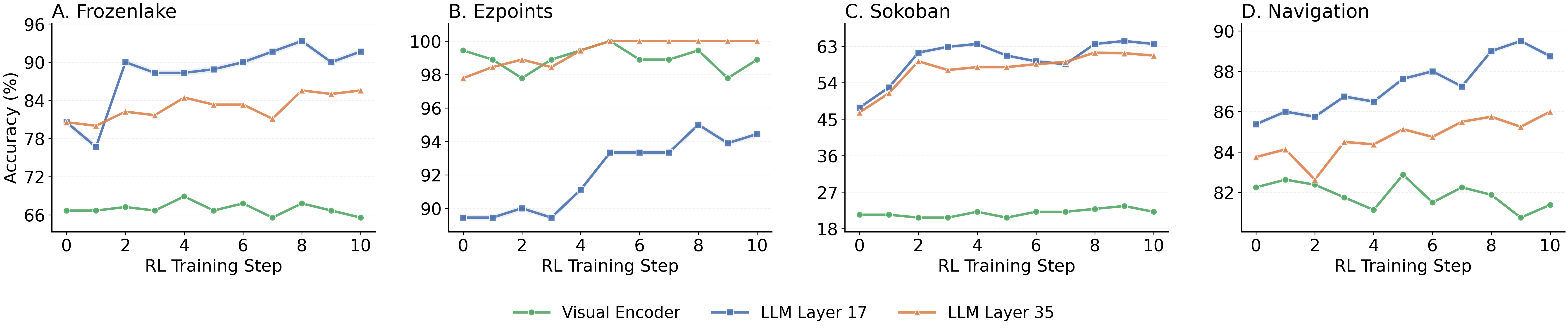}
  \caption{Probe accuracy of the visual encoder and two LLM layers across RL training checkpoints
  in FrozenLake, EzPoints, Sokoban, and Navigation.
  The visual encoder (green) remains flat throughout training, while LLM Layer 17 (blue)
  and LLM Layer 35 (red) improve steadily, with most gains concentrated in the early checkpoints.}
  \label{fig:probe}
\end{figure*}

In the preceding experiments, we observed that multi-turn RL significantly improves
the grounded VQA capabilities of VLMs in these settings.
To examine factors associated with this improvement,
we use representational probing to track the internal representation space
across 11 checkpoints throughout RL optimization.
This section analyzes how the vision encoder and LLM backbone change during training,
and uses attention weight analysis to investigate a contributing mechanism
of grounded VQA improvements under RL.

\subsection{Visual Inertia and LLM Representational Reshaping}
\label{subsec:probing_results}

We first investigate the capacity of the vision encoder to extract low-level environmental state information.
Probing results across all 11 checkpoints reveal a counter-intuitive phenomenon:
vision-encoder probe accuracy remains relatively stable throughout RL training,
as illustrated in Figure~\ref{fig:probe}.

Specifically, in the FrozenLake environment,
the probe accuracy of visual features fluctuates within the narrow band of 65.56\% to 68.89\%,
showing no discernible upward trend with training steps.
In the Sokoban environment, which imposes substantially higher planning complexity,
the probe accuracy equally stagnates between 20.71\% and 23.57\%, with a similar flat trend observed in Navigation.
The sole exception is the EzPoints environment,
where the probe accuracy of the vision encoder remains stably near perfect throughout training
(97.78\%--100.00\%).
This ceiling effect indicates that task-relevant card values are already encoded
in the low-level visual features from the outset.
These observations show no consistent improvement in the vision encoder's decodability
on our diagnostic tasks during multi-turn interaction and policy optimization.
Stable probe accuracy does not exclude representation drift.

In contrast, the LLM's internal representations change substantially throughout RL training.
We specifically track the intermediate layer (layer 17) and the deep layer (layer 35) of the LLM.
In the FrozenLake task,
the probe accuracy of layer 17 rises from 80.56\% at initialization 
to 93.33\% at checkpoint~8, an absolute gain of $+$12.77\%,
subsequently stabilizing at 91.67\%;
layer 35 concurrently improves from 80.56\% to 85.56\%.
This improvement is even more pronounced in the Sokoban environment:
layer 17 increases from 47.86\% to 63.57\% ($+$15.71\%),
while layer 35 rises from 46.43\% to 60.71\% ($+$14.29\%).
The Navigation environment mirrors this upward trajectory in the deep layers.
Notably, the majority of these representational gains in the deep LLM layers
are concentrated within the first three RL checkpoints.
This rapid early-stage shift shows that RL quickly reshapes
the LLM's internal representations,
prompting the model to encode game states more relevant to spatial decision-making.

Additional RL experiments on Qwen3-VL-4B in FrozenLake show the same probing pattern:
vision-encoder probe accuracy remains relatively stable, while LLM-layer probe accuracy improves.
Grounded VQA accuracy also increases from $42.22\%$ to $51.89\%$.
Full checkpoint results are reported in Appendix~\ref{sec:appendix_qwen3_rl}.

\subsection{Core Insight: Visual Utilization as a Contributing Factor}
\label{subsec:core_insight}
\begin{figure}[t]
 
    \includegraphics[width=\linewidth]{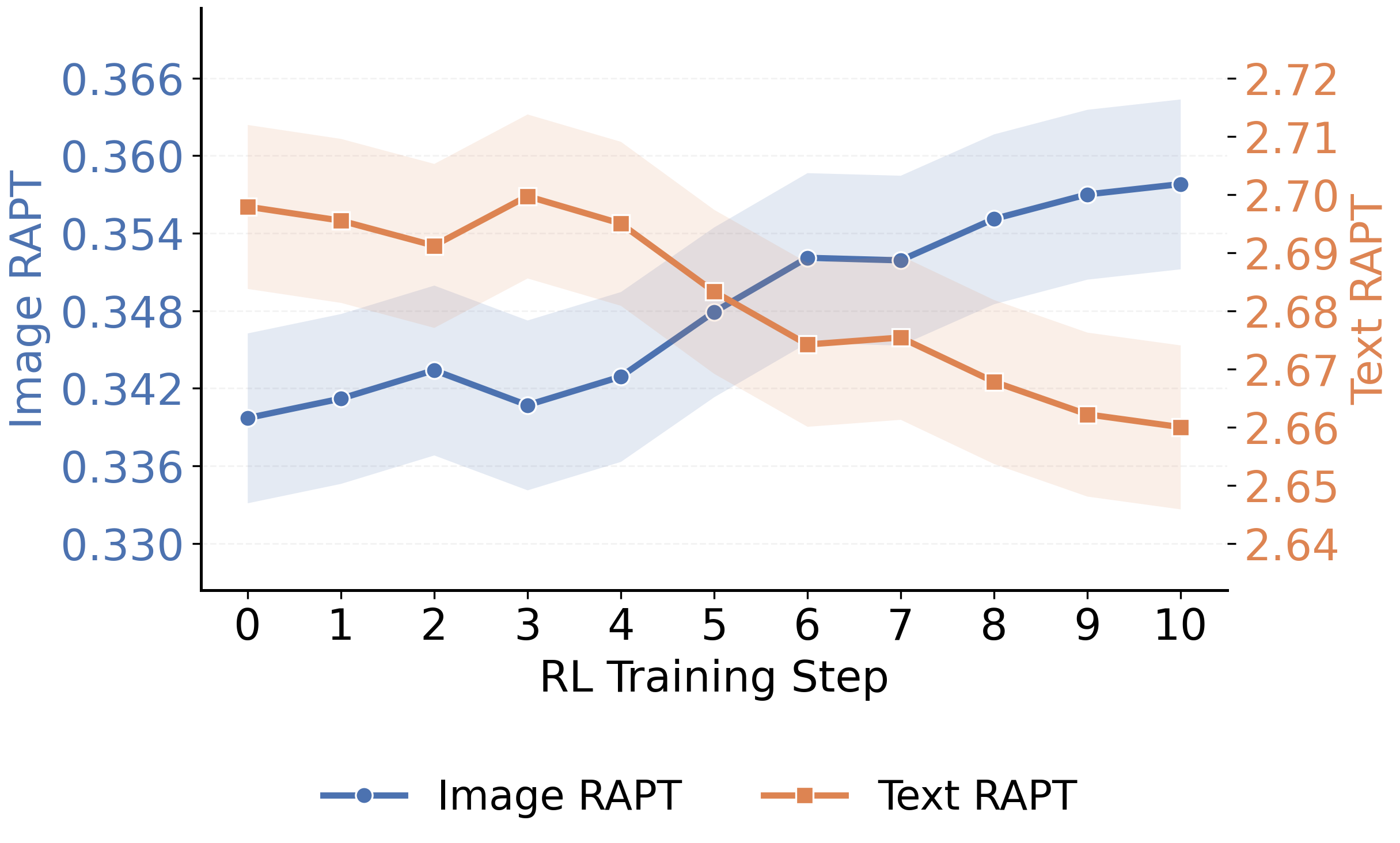}
  
  \hfill
 
    \caption{Evolution of image RAPT (blue, left axis) and text RAPT (red, right axis)
    across RL training checkpoints.
    Image RAPT rises steadily while text RAPT declines, indicating that RL systematically
    shifts attention from text toward visual tokens.}
    \label{fig:rapt}
 
\end{figure}

Together, the probing results suggest that RL improves the LLM's utilization
of visual information already present in its activations~\cite{fu2025hiddenplainsightvlms}.

To understand how the LLM improves its visual utilization,
we turn to the transformer's attention mechanism.
We use the Relative Attention per Token (RAPT) metric~\cite{ref10} for quantitative analysis.
RAPT is the ratio of the average attention weight received by tokens of a specific modality
(visual or textual) to the global per-token average across the full input.
This normalizes for token-count imbalance in long contexts and captures
the model's cross-modal attention preference at the token level.
We compute layer-wise RAPT values for image tokens at each RL checkpoint,
producing RAPT evolution curves stratified by training stage, as shown in Figure~\ref{fig:rapt}.

The curves show a consistent pattern across environments.
In the base model initialization, the attention distribution in deep layers exhibits a strong textual prior:
the RAPT curve for visual tokens remains low,
which explains why the initial model fails to exploit
the visual representations already present in its activations.
As RL training progresses,
the visual token RAPT in deep layers rises steadily and persistently.
This redistribution of attention provides clear mechanistic evidence that
RL reduces the model's textual bias, directing more attention toward visual content,
with the shift most pronounced in the deep layers of the LLM.

\paragraph{Persistent Textual Prior and Marginal Attention Shift.}
We analyze the evolution of RAPT values for both image and text modalities
across all 11 checkpoints.
At initialization (checkpoint~0), the text RAPT is $2.6979$,
approximately eight times the image RAPT of $0.3397$,
confirming a pronounced textual prior in the baseline model.
As RL progresses, we observe a continuous but modest redistribution.
By checkpoint~10, the text RAPT declines to $2.6600$,
while the image RAPT rises to $0.3578$.
Although the shift is small in absolute terms,
the inverse trajectories of the two modalities confirm that RL systematically
reallocates attention from text toward visual features.

\paragraph{Mechanistic Interpretation: Small Attention Shifts and Grounded VQA Gains.}
The RAPT trends are consistent with attention reallocation contributing to the performance gains.
The model does not undergo a complete attention inversion; it retains its language prior.
Instead, it learns to consistently increase its use of visual state information.
The image RAPT grows by only $0.0181$ over the full RL trajectory,
and this shift is associated with improved grounded VQA performance.
The probe and attention results suggest that limited utilization of visual information
may be a bottleneck in our settings, with attention reallocation under RL
contributing to improved grounded VQA performance.

\begin{table*}[t]
  \centering
  \caption{Ablation study of the fixed-bias attention intervention on the base model.
  Accuracy (\%) is reported for six configurations across four environments;
  the best result in each column is in bold.}
  \label{tab:ablation}
  \begin{tabular}{lcccc}
    \toprule
    Configuration & FrozenLake & EzPoints & Sokoban & Navigation \\
    \midrule
    No intervention          & 28.89 & 82.22 & 25.00 & 78.75 \\
    Image token + deep layers & \textbf{33.89} & \textbf{83.33} & \textbf{27.86} & 80.50 \\
    \midrule
    Text token + deep layers  & 23.33 & 77.78 & 24.29 & 78.75 \\
    All prompt tokens + deep layers  & 30.56 & 81.67 & 26.43 & 78.88 \\
    \midrule
    Image token + shallow layers & 31.67 & 81.67 & 26.43 & 79.00 \\
    Image token + all layers     & \textbf{33.89} & 82.22 & 26.43 & \textbf{80.63} \\
    \bottomrule
  \end{tabular}
\end{table*}

\section{Intervention}
\label{sec:intervention}

\begin{figure}[t]
  
    \includegraphics[width=\linewidth]{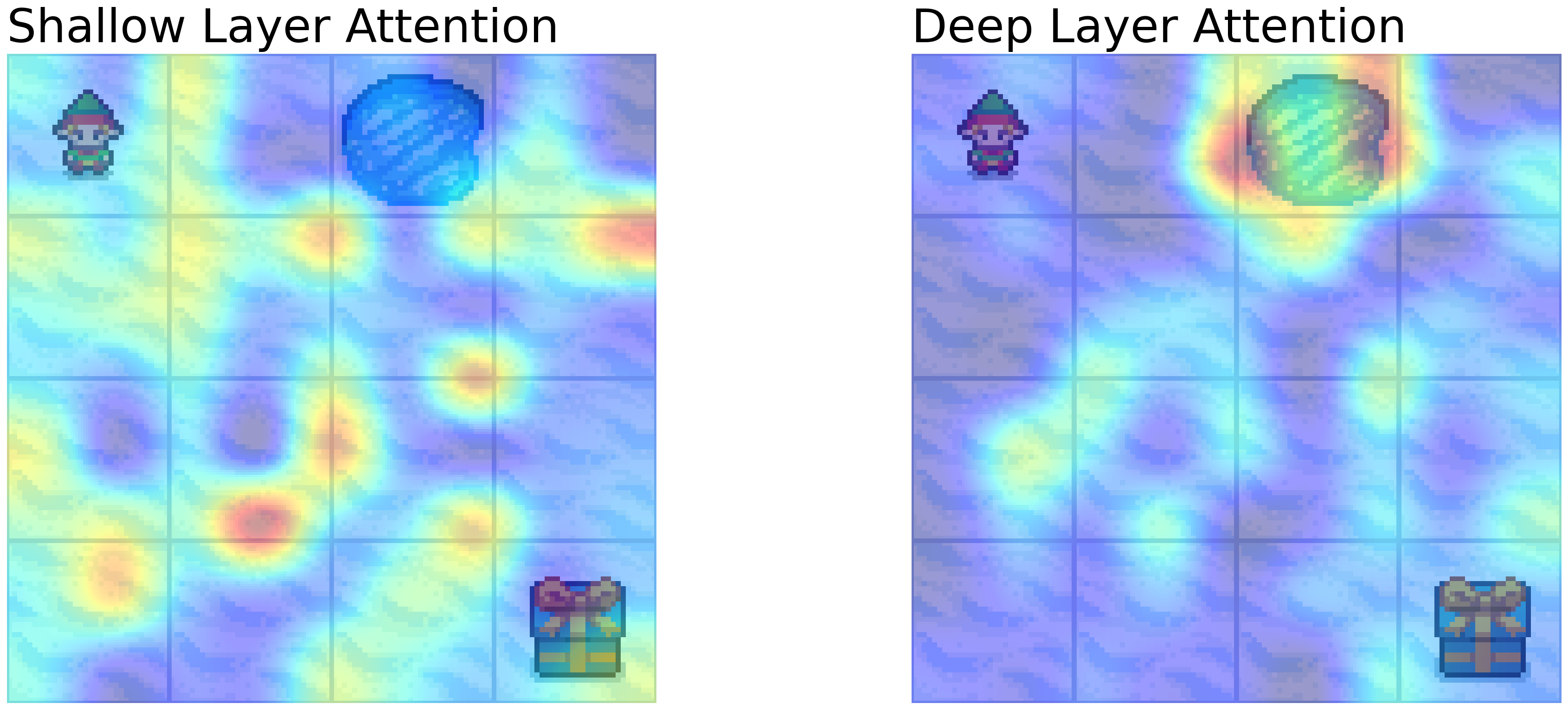}
 
  \hfill
  
    \caption{Attention heatmaps for shallow (left) and deep (right) layers in FrozenLake.
    Shallow layers distribute attention diffusely across the grid, while deep layers
    concentrate on task-relevant objects (the hole), consistent with the layer-wise
    attention shift observed during RL training.}
    \label{fig:attn_layers}
  
\end{figure}

\subsection{Motivation}
\label{subsec:motivation}

The mechanistic analysis in Section~\ref{sec:mechanism} points to a clear hypothesis:
if the bottleneck to grounded visual understanding lies in insufficient attention to image tokens,
can an inference-time intervention on the attention distribution partially reproduce
the performance gains achieved by full RL fine-tuning?
To test this, we design a lightweight, fixed-bias attention module that increases the attention
allocated to visual tokens in the deep layers of the LLM.
This intervention operates entirely at inference time, requiring no parameter updates,
no gradient computation, and no environment interaction.
Applying the intervention to the base model allows us to test whether changing attention allocation alone can improve grounded VQA performance.

\subsection{Method Formulation}
\label{subsec:formulation}

Formally, let the standard scaled dot-product attention be defined as:
\begin{equation}
  \mathrm{Attn}(Q, K, V) = \mathrm{Softmax}\!\left(\frac{QK^{\top}}{\sqrt{d_k}}\right)V.
\end{equation}
We introduce an additive visual attention bias $\lambda \cdot M$ into the pre-softmax
logits, yielding the modified attention computation:
\begin{equation}
  \mathrm{Attn}'(Q, K, V) =
    \mathrm{Softmax}\!\left(\frac{QK^{\top}}{\sqrt{d_k}} + \lambda \cdot M\right)V,
  \label{eq:intervention}
\end{equation}
where $\lambda > 0$ is a scalar controlling the strength of the visual bias,
and $M \in \{0, 1\}^{L}$ is a binary modality mask over the full input sequence of length $L$:
each position corresponding to an image token is assigned a value of $1$,
while each position corresponding to a text token is assigned $0$.
The bias increases visual attention without gradient computation or model parameter updates.
An analysis using a trained bias is provided in Appendix~\ref{sec:appendix_trained_intervention}.

Adding the bias before softmax increases the relative weight of visual keys while preserving their ordering within the visual modality. This intervention adjusts how existing visual information is weighted without introducing new visual features.

As illustrated in Figure~\ref{fig:attn_layers},
shallow Transformer layers in FrozenLake exhibit a pronounced textual prior,
consistent with the phenomenon reported in prior work~\cite{ref10,zhang2025mllms},
whereas deeper layers show a relative shift toward visual content.
This layer-wise pattern motivates applying the bias intervention
defined in Equation~\ref{eq:intervention} exclusively to the deep layers of the LLM backbone.
Applying it to shallow layers risks disrupting the low-level linguistic processing
on which the model relies for syntactic and semantic grounding.

\subsection{Ablation Study}
\label{subsec:ablation}

\paragraph{Image Tokens Are the Optimal Intervention Target.}
Among configurations applied to deep layers, biasing attention toward image tokens alone
yields the highest accuracy across the evaluated environments:
$33.89\%$, $83.33\%$, $27.86\%$, and $80.50\%$ on FrozenLake, EzPoints, Sokoban, and Navigation,
gains of $5.00$, $1.11$, $2.86$, and $1.75$ percentage points over the no-intervention baseline.
Extending the bias to all prompt tokens reduces performance relative to the image-only condition,
indicating that biasing text positions partially offsets the gain.
Biasing text tokens alone is more damaging still:
accuracy drops below or stagnates at the no-intervention baseline, falling to $23.33\%$ on FrozenLake and $77.78\%$ on EzPoints.
This confirms that reinforcing the textual prior that RL suppresses is directly harmful
to grounded visual understanding.

\paragraph{Deep Layers Provide the Optimal Intervention Site.}
Among image-token configurations, restricting the bias to deep layers outperforms
the shallow-layer counterpart on all evaluated benchmarks:
$33.89\%$ vs.\ $31.67\%$ on FrozenLake,
$83.33\%$ vs.\ $81.67\%$ on EzPoints, $27.86\%$ vs.\ $26.43\%$ on Sokoban, and $80.50\%$ vs.\ $79.00\%$ on Navigation.
Applying the bias across all layers also generally fails to match the deep-only result,
despite covering a broader portion of the network:
it ties on FrozenLake, yields only a marginal gain on Navigation ($80.63\%$ vs.\ $80.50\%$), but falls short on EzPoints ($82.22\%$ vs.\ $83.33\%$) and Sokoban ($26.43\%$ vs.\ $27.86\%$).
This is consistent with the layer-wise analysis in Figure~\ref{fig:attn_layers}:
shallow layers handle low-level linguistic processing,
and adding a visual bias there introduces a competing signal
that interferes with syntactic and semantic grounding.
Deep-layer intervention is therefore both sufficient and preferable,
targeting the stage where cross-modal routing decisions are made
while leaving early-layer representations intact.

\section{Conclusion}
Across four diverse agent environments, we find that multi-turn RL fine-tuning improves grounded VQA performance in VLM agents, outperforming SFT in our evaluations. Representational probing shows that vision-encoder probe accuracy remains relatively stable, while LLM-layer probe accuracy improves. Attention analysis further indicates increased attention to visual tokens during RL training.

This insight inspired a lightweight, inference-time attention bias plugin that elevates visual token attention without modifying any model parameters. Our results support visual information utilization in the LLM backbone as a contributing mechanism of grounded VQA gains in the evaluated settings.
These findings suggest that future RL methods for VLM agents should explicitly consider how the language backbone uses visual information.

\section*{Limitations}

While our findings support attention reallocation as a contributing mechanism of grounded VQA improvements under RL, bridging these gains with end-to-end multi-step task success remains an open direction.
Successfully completing an embodied task requires both correct perception of the environment and effective sequential planning. While our fixed-bias attention plugin provides interventional evidence for the attention-shift hypothesis, it does not explicitly capture the multi-step reasoning and spatial policy knowledge acquired by the actor during RL fine-tuning. Thus, inference-time interventions may best serve as a complement to, rather than a full replacement for, the end-to-end RL pipeline in complex decision-making.
In addition, although we examine multiple RL checkpoints, a finer-grained analysis of the very early training dynamics may yield deeper insights into exactly how this attention reallocation emerges.

\bibliography{custom}

\appendix

\section{Selection of Probed Layers}
\label{sec:appendix_layer_selection}

In Section~\ref{sec:mechanism}, we specifically track the representation spaces of Layer 17 and Layer 35 in the LLM backbone. The Qwen2.5-VL-3B model consists of 36 transformer layers. We selected Layer 17 as it represents the approximate mid-point of the network, capturing the intermediate state of cross-modal fusion after the shallow layers' initial syntactic and semantic processing. Layer 35, being the final transformer layer, captures the final, fully-contextualized representation state immediately prior to the language modeling head for decoding. Tracking these two specific layers allows us to observe the evolution of visual representations from deep intermediate reasoning to the final output distributions.

\section{Hyperparameter Selection for Fixed-Bias Intervention}
\label{sec:appendix_hyper_search}

In Section~\ref{sec:intervention}, we introduced an additive visual attention bias to elevate the attention allocated to image tokens. For the fixed-bias variant evaluated in the ablation study, the optimal scalar controlling the strength of the visual bias (denoted as $\gamma$, corresponding to the $\lambda$ scalar in Equation~\ref{eq:intervention}) was determined via a grid search over the candidate values $\{0.0, 0.25, 0.5, 0.75, 1.0, 1.5, 2.0, 3.0, 5.0\}$. The bias was selected using five-fold cross-validation and fixed for inference-time evaluation.

In the all-prompt-token ablation, the bias applies to the original image and text tokens;
newly generated tokens receive zero bias during autoregressive decoding.

\section{Trained Attention Bias Validation}
\label{sec:appendix_trained_intervention}

To further test the attention-shift hypothesis and whether the bias can be optimized, we evaluate an alternative variant: the \textit{trained plugin}. In this variant, $\lambda$ is optimized on a small VQA training set with all primary model parameters frozen.

\subsection{Accuracy Gains over RL Baseline}

\begin{figure*}[h]
  \centering
  \includegraphics[width=\linewidth]{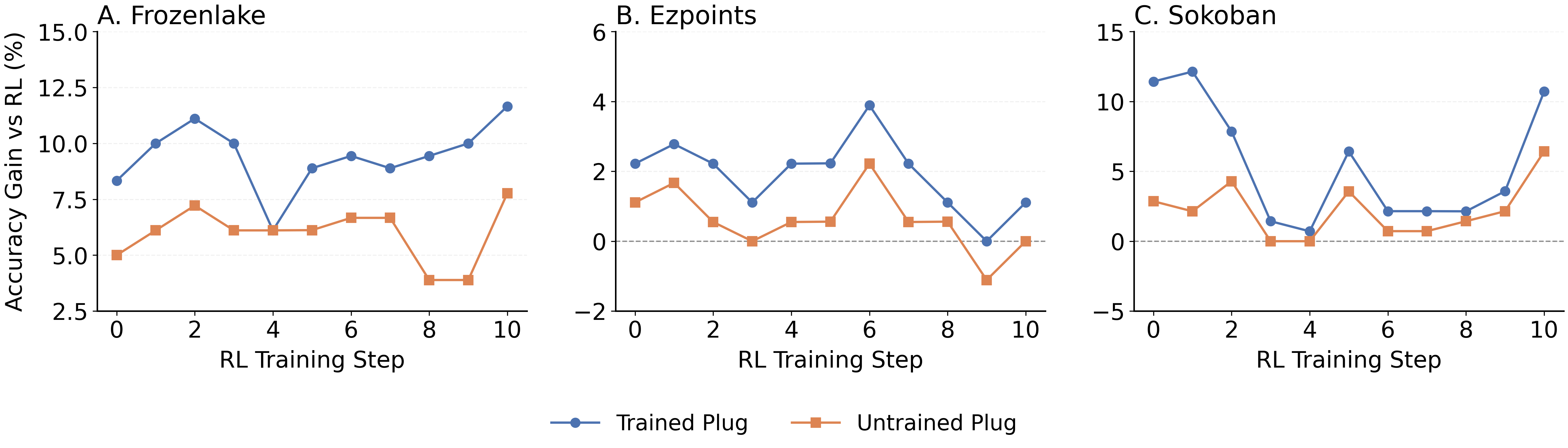}
  \caption{Accuracy gain of the trained plugin (blue) and fixed-bias variant (red) over the RL baseline
  across checkpoints in FrozenLake, EzPoints, and Sokoban.
  The trained plugin consistently achieves larger gains, while the fixed-bias variant shows higher
  variance and occasionally falls below the RL baseline, confirming that the learned bias
  values are key to reliable improvement.}
  \label{fig:intervention}
\end{figure*}

As shown in Figure~\ref{fig:intervention}, both variants produce positive accuracy gains at nearly every checkpoint. In FrozenLake, the trained plugin yields gains of approximately $6.1$--$11.7$ percentage points over the RL baseline, while the fixed-bias variant achieves $3.9$--$7.8$ points. In Sokoban, the gap is particularly large at early checkpoints (e.g., $+12.1$ vs.\ $+2.1$ at checkpoint~1), suggesting that the trained plugin learns a task-relevant attention pattern that the fixed-$\lambda$ variant cannot replicate.

\subsection{Plugin Training Validates the RL Attention Shift Direction}
\label{subsubsec:plugin_rapt}

\begin{figure}[h]
 
    \includegraphics[width=\linewidth]{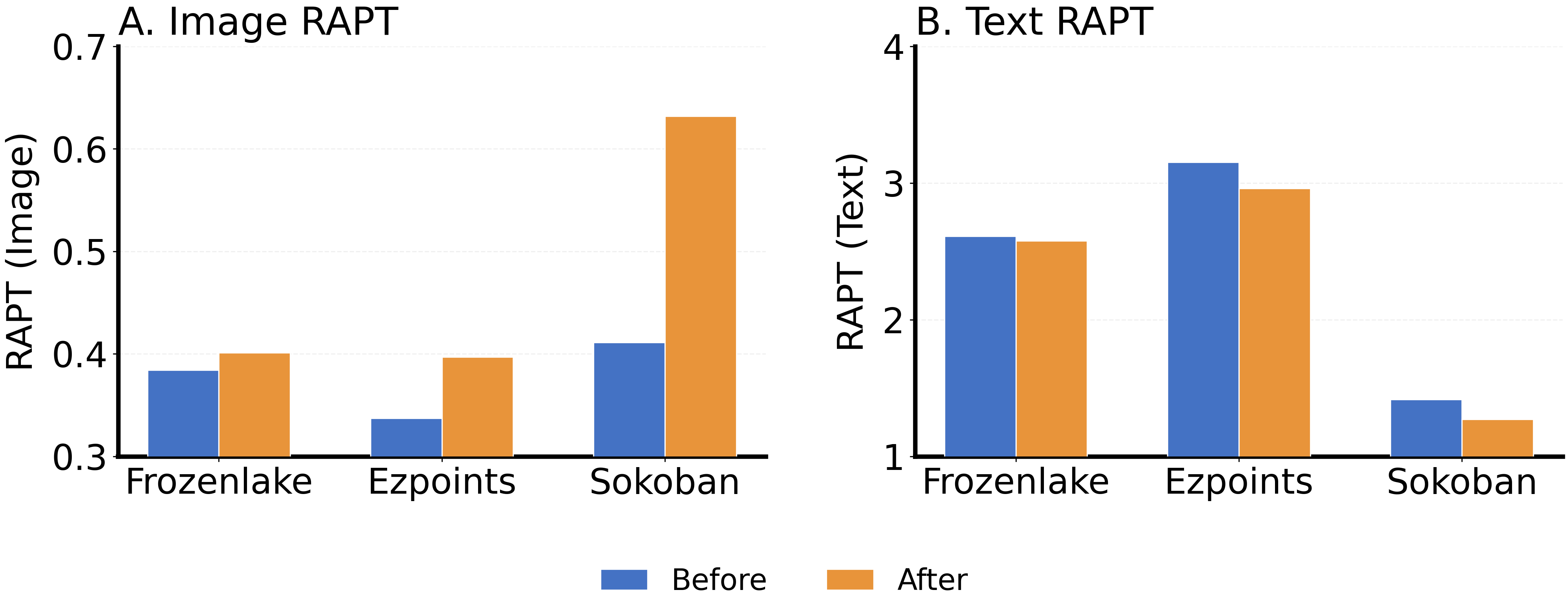}

  \hfill
  
    \caption{Image RAPT (A) and text RAPT (B) before and after plugin training across
    three environments. Plugin optimization consistently increases image RAPT
    while decreasing text RAPT in all three environments, mirroring the attention
    redistribution direction induced by full RL fine-tuning.}
    \label{fig:train_rapt}
  
\end{figure}

To confirm that the plugin learns a direction consistent with RL-induced redistribution, we initialize $\lambda = 0$ and record RAPT before and after optimizing $\lambda$. As shown in Figure~\ref{fig:train_rapt}, plugin training consistently raises image RAPT while lowering text RAPT across all three environments (e.g., in Sokoban, image RAPT rises from $0.411$ to $0.632$ while text RAPT falls from $1.417$ to $1.272$). This inverse co-movement, arising solely from gradient-based optimization on VQA, is consistent with the attention-shift hypothesis for grounded visual understanding.

\section{Use of LLMs}

In this study, Large Language Models (LLMs) were employed solely in auxiliary roles. First, they supported the preparation of the manuscript by assisting with grammatical corrections, language refinement, and improvements in academic clarity. Second, during execution, their deep research capabilities were utilized to obtain a broader and more comprehensive overview of related work. These uses were strictly supportive in nature and did not influence the design, implementation, or analysis of the proposed pipeline.


\begin{table*}[t]
  \centering
  \caption{RL (GRPO) hyperparameters.}
  \label{tab:rl_hparams}
  \begin{tabular}{lccc}
    \toprule
    \textbf{Hyperparameter} & \textbf{FrozenLake} & \textbf{EzPoints} & \textbf{Sokoban} \\
    \midrule
    Total training steps             & 150              & 150              & 150              \\
    Actor learning rate              & $1\times10^{-6}$ & $1\times10^{-6}$ & $1\times10^{-6}$ \\
    Train batch size                 & 16               & 16               & 32               \\
    PPO mini-batch size              & 32               & 32               & 32               \\
    Max prompt length (tokens)       & 1{,}024          & 1{,}024          & 1{,}024          \\
    Max response length (tokens)     & 200              & 256              &  512             \\
    Max trajectory length (tokens)   & 2{,}400          & 5{,}120          & --               \\
    Max turns per episode            & 3                & 5                & map-dependent    \\
    Rollout temperature              & 0.7              & 0.7              & 0.7              \\
    Top-$p$                          & 0.95             & 0.95             & 0.95             \\
    KL coefficient                   & 0.001            & 0.001            & 0.001            \\
    \texttt{use\_kl\_loss}           & False            & False            & False            \\
    \texttt{use\_multi\_turn\_reward}& False            & True             & False            \\
    GPUs                             & 2                & 2                & 2                \\
    Training set size                & 10{,}000         & 10{,}000         & 10{,}000         \\
    Test set size                    & 128              & 128              & 128              \\
    \bottomrule
  \end{tabular}
\end{table*}

\begin{table}[t]
\centering
\small
\setlength{\tabcolsep}{6pt}
\caption{VQA accuracy (\%) of the attention bias plugin on Qwen3-VL, LLaVA, and Gemma across FrozenLake, EzPoints, and Sokoban.}
\label{tab:arch_intervention}
\begin{tabular}{lccc}
\toprule
Architecture & FrozenLake & EzPoints & Sokoban \\
\midrule
Qwen3VL-O & 42.22 & 90.56 & 37.86 \\
Qwen3VL-P & 53.33 & 92.22 & 39.29 \\
LLaVA-O   & 41.67 & 73.33 & 22.86 \\
LLaVA-P   & 44.44 & 73.89 & 24.29 \\
Gemma-O   & 24.44 & 90.00 & 10.71 \\
Gemma-P   & 27.22 & 91.67 & 13.57 \\
\bottomrule
\end{tabular}
\end{table}

\section{Dataset Construction Details}
\label{sec:appendix_dataset}

\subsection{VQA Dataset}
\label{subsec:appendix_vqa}

We construct separate VQA datasets for each environment to train visual perception probes that assess whether models can extract key information from observations. Representative samples are shown in Figure~\ref{fig:env_examples}.

\paragraph{FrozenLake.}
The FrozenLake dataset comprises 900 RGB screenshots of $4\times4$ grid maps (Gymnasium \texttt{FrozenLake-v1}), drawn uniformly across nine hole-count categories (0--8 holes, 100 images each), with ice-tile probability sampled from $p \in [0.4, 0.95]$. Each sample asks \textit{``How many holes are in this map? Answer with only the number.''}, with the hole count (0--8) as the label; each record also stores pixel coordinates of all holes for subsequent analysis.

\paragraph{EzPoints.}
The EzPoints dataset also contains 900 samples (100 per class). Each screenshot shows two cards under full-frame Gaussian blur (radius $= 2.0$), with one highlighted by a red circle; the question asks \textit{``What is the numeric value of the card in the red circle? Answer with only the number.''} with labels in $[2, 10]$. The blur simulates visual degradation in late-stage RL training, and the red circle designates the query target, decoupling perceptual difficulty from arithmetic reasoning.

\paragraph{Sokoban.}
The Sokoban dataset covers seven box-count categories (1--7 boxes, 100 images each, 700 total), with images randomly generated and rendered as RGB screenshots. Map size and generation steps increase with box count to handle more complex layouts. The question follows the fixed template: \textit{``This is a Sokoban puzzle. Answer the question based on the image.''}

\paragraph{Navigation.}
The Navigation dataset comprises 800 samples collected from the AI2-THOR simulator~\cite{kolve2017ai2}.
Images are egocentric first-person RGB camera views from 54 indoor scenes.
For each sample, the question asks: \textit{``What object do you see directly in front of you? Choose from: GarbageCan, Laptop, Pot, DeskLamp, Toaster, AlarmClock, Box, Cup, Pillow''}.
The label corresponds to the ground-truth object directly ahead of the agent.

All datasets are stored in \texttt{.parquet} format compatible with the HuggingFace \texttt{datasets} library, and partitioned into training and test sets at a 4:1 ratio with a fixed random seed (\texttt{random\_state=42}).

\subsection{SFT Dataset}
\label{subsec:appendix_sft_data}

The SFT dataset trains the model to generate expert actions with explicit reasoning in the format \texttt{<think>...\allowbreak</think>}\allowbreak\texttt{<answer>...\allowbreak</answer>}.

\paragraph{FrozenLake.}
For each seed, BFS computes the shortest hole-free path to the goal. At each step, the observation is saved as a PNG; a reasoning trace is generated from the agent's position, goal, and nearby obstacles (holes, walls, goal tile); and the trace with the chosen action are stored as \texttt{<think>\{reasoning\}\allowbreak</think>}\allowbreak\texttt{<answer>\{action\}\allowbreak</answer>}.

\paragraph{EzPoints.}
Each EzPoints episode consists of a fixed four-step sequence: select the first operand, select the operator ($+$ or $\times$), select the second operand, and submit the equality token $=$. The expert policy applies addition when $c_1 + c_2 = \text{target}$ and multiplication otherwise. A stage-specific reasoning template is generated at each step, and the output follows the same \texttt{<think>...\allowbreak</think>}\allowbreak\texttt{<answer>...\allowbreak</answer>} format.

\paragraph{Sokoban.}
For each seed, BFS computes the shortest path to the target configuration. At each step, the state is saved as a PNG; a reasoning trace is generated from the player's position, target locations, and nearby obstacles (walls, boxes, target tiles); and the trace with the selected action are formatted as \texttt{<think>\{reasoning\}\allowbreak</think>}\allowbreak\texttt{<answer>\{action\}\allowbreak</answer>}.

\paragraph{Navigation.}
For each ObjectNav task, expert trajectories are generated using a rule-based scheme computing the shortest path to the target. At each step, a detailed trace is formulated reflecting spatial grounding, structured as: \texttt{<think>\{reasoning\}</think><answer>\{actions\}
</answer>}.

All SFT datasets are loaded through the VAGEN environment interface with \texttt{render\_mode=vision} and \texttt{prompt\_format=free\_think}, and exported as JSON files compatible with fine-tuning frameworks such as LLaMA-Factory.

\section{Training Details}
\label{sec:appendix_impl}

\subsection{Supervised Fine-Tuning (SFT)}
\label{subsec:appendix_sft}

We trained Qwen2.5-VL-3B with full-parameter fine-tuning using LLaMA-Factory
and DeepSpeed ZeRO Stage~3 in bf16; full hyperparameters are in Table~\ref{tab:sft_hparams}.

\begin{table}[h]
  \centering
  \caption{SFT hyperparameters.}
  \label{tab:sft_hparams}
  \begin{tabular}{lc}
    \toprule
    \textbf{Hyperparameter} & \textbf{Value} \\
    \midrule
    Training epochs       & 4 \\
    Learning rate         & $1\times10^{-5}$ \\
    LR schedule           & cosine \\
    Warmup ratio          & 0.1 \\
    Gradient accumulation & 2 \\
    Max sequence length   & 4{,}096 \\
    Precision             & bf16 \\
    DeepSpeed stage       & ZeRO-3 \\
    \bottomrule
  \end{tabular}
\end{table}

\subsection{Reinforcement Learning (RL)}
\label{subsec:appendix_rl}

We trained GRPO policies via VAGEN and verl-agent (both built on VeRL) with a masked advantage
estimator, initialising the actor from the same Qwen2.5-VL-3B base weights used for SFT. Rollouts were generated with
vLLM (tensor-parallel degree~2) at temperature 0.7 and top-$p$ 0.95, with both loss
masking (\texttt{use\_loss\_mask}) and advantage masking (\texttt{use\_gae\_mask}) enabled
to suppress gradients from padding and non-decision tokens.

\paragraph{FrozenLake.}
The agent navigates a grid-world from start to goal, avoiding holes under slippery-tile
dynamics. Each turn it receives a rendered top-down image of the grid.
The full system prompt reads:
\begin{promptbox}{FrozenLake Prompt}
You are a FrozenLake solver.

\textbf{FrozenLake Quick Guide}

Goal: Reach the goal (G).

Symbols (If image is provided there are no symbols):\\
\texttt{\_}~Frozen $|$ \texttt{O}~Hole $|$ \texttt{G}~Goal $|$ \texttt{P}~Player $|$ \texttt{X}~Player fell into hole $|$ \ensuremath{\checkmark}~Player on goal

Rules:\\
1.~Avoid falling into holes.\\
2.~Frozen tiles are slippery, you may move perpendicular to your intended direction.

Actions you can take: Left, Down, Right, Up.

You can take up to \texttt{\{max\_actions\_per\_step\}} action(s) at a time, separated by~`,'. You should first give your reasoning, and then your answer. Your response should be in the format of \texttt{<think>}\ldots\texttt{</think>}\texttt{<answer>}\ldots\texttt{</answer>}.

e.g.\ \texttt{<think>}I can see the target is on my down left, I should go down then left to reach the target\texttt{</think><answer>Down,Left</answer>}
\end{promptbox}
Each episode runs for at most 3 turns.  The agent outputs a comma-separated action
sequence per turn, e.g.\ \texttt{<answer>Down,Left</answer>}.  The reward is binary:
$+1$ on successfully reaching the goal, $0$ otherwise.  Responses that do not conform
to the required \texttt{<think>}/\texttt{<answer>} format incur a penalty of $-0.02$.
Training ran for 150 GRPO steps with per-step multi-turn reward disabled (reward
assigned only at trajectory end, \texttt{use\_multi\_turn\_reward=False}).

\paragraph{EzPoints.}
The environment is a card-arithmetic game: the agent is dealt two number cards and must
use them with operators $\{+,\times\}$ to equal a target value (12 by default).
The agent takes one action per turn (picking a number, an operator, or submitting~\texttt{=}),
with up to 5 turns per episode.  The full system prompt reads:
\begin{promptbox}{EzPoints Prompt}
You are playing EZPoints.

Goal: Use the two cards shown in the image and operators (+, *) to get exactly~12.

Rules:\\
1.~Each turn, you can only take EXACTLY ONE action (a number, an operator, or `=').\\
2.~You must use each card exactly once.\\
3.~Available operators: `+', `*'\\
4.~Use `=' after you have constructed the full formula to evaluate it.

Example Move Sequence:\\
Turn~1: \texttt{<think>}I will pick 6.\texttt{</think><answer>6</answer>}\\
Turn~2: \texttt{<think>}I will pick *.\texttt{</think><answer>*</answer>}\\
Turn~3: \texttt{<think>}I will pick 2.\texttt{</think><answer>2</answer>}\\
Turn~4: \texttt{<think>}Now I evaluate 6*2.\texttt{</think><answer>=</answer>}

You can take up to 1 action(s) at a time, separated by~`,'. You should first give your reasoning, and then your answer. Your response should be in the format of \texttt{<think>}\ldots\texttt{</think>}\texttt{<answer>}\ldots\texttt{</answer>}.

e.g.\ \texttt{<think>}I have a 6 and a 2, and the target is 12. 6 times 2 is 12.\texttt{</think><answer>6</answer>}
\end{promptbox}
The reward signal is multi-turn: a correctness reward is given when the submitted
expression evaluates to the target, invalid actions incur a penalty of $-0.1$, and a
format reward of $+0.1$ is granted for well-formed responses.  Training ran for 150
GRPO steps with \texttt{use\_multi\_turn\_reward=True} to propagate rewards at each
decision step.

\paragraph{Sokoban.}
The agent operates in a $6\times6$ grid-world, pushing boxes onto target spots.
Since boxes can only be pushed (not pulled), trapping one in a corner fails the episode,
requiring long-horizon planning. The agent selects one action per turn from
\{\texttt{up}, \texttt{down}, \texttt{left}, \texttt{right}\},
wrapped in \texttt{<action>...</action>} tags. The full system prompt reads:
\begin{promptbox}[left=3mm,right=3mm]{Sokoban Prompt}
You are an expert agent operating in the Sokoban environment. Your goal is to push all the boxes onto the target spots. Once all boxes are on the targets, you win!

\textbf{\# Rules}

You can only push boxes. You can't pull them, so plan ahead to avoid getting stuck.\\
You can't walk through or push boxes into walls.\\
To avoid traps, do not push boxes into corners or against walls where they can't be moved again.

\textbf{\# Visual Elements in the Image:}

Character: A small, green alien-like figure with two antennae and black eyes. It represents you.\\
Box: A yellow crate marked with an orange ``X'' across its front. It is the box you need to push.\\
Target: A black tile outlined in red, with a small red diamond shape in the center. It marks the destination where a box should be pushed.

\textbf{\# Current Step}

Your current observation is shown in the image.\\
Your admissible actions are [\texttt{up}, \texttt{down}, \texttt{left}, \texttt{right}].\\
Now it's your turn to make a move (choose ONE action only for the current step).\\
You should first reason step-by-step about the current situation --- observe the positions of boxes and targets, plan a path to push a box toward a target, and avoid traps like corners or walls. This reasoning process MUST be enclosed within \texttt{<think>} tags.\\
Once you've finished your reasoning, you should choose an admissible action for the current step and present it within \texttt{<action>} tags.
\end{promptbox}

Episode length is map-dependent. The reward structure mirrors EzPoints: $+10$ for
placing all boxes on targets, $-0.1$ for invalid actions (e.g., pushing into a wall),
and a format reward for well-formed \texttt{<think>}/\texttt{<action>} tags.
Reward is assigned at trajectory end (\texttt{use\_multi\_turn\_reward=False}).

\paragraph{Navigation.}
The AI2-THOR ObjectNav environment requires the agent to find specific objects
within immersive 3D indoor scenes. The agent executes sequences of up to 5 actions per turn from an 8-action vocabulary.
The full system prompt reads:
\begin{promptbox}{Navigation Prompt}
You are a home robot and perform navigation tasks according to instructions.
Actions you can take: moveahead, moveback, moveright, moveleft, rotateright, rotateleft, lookup, lookdown. 
moveahead: Move forward by some distance
moveback: Move backward by some distance
moveright: Move rightward by some distance
moveleft: Move leftward by some distance
rotateright: Rotate to the right by 90 degrees
rotateleft: Rotate to the left by 90 degrees
lookup: Tilt the camera upward by 30 degrees
lookdown: Tilt the camera downward by 30 degrees
Rewards:
Format correct: +0.5
Achieve the human instruction: +10.0
The instruction will be provided with each observation. Look at the image carefully and navigate to complete the instruction.
Hints:
1. You can take multiple actions at a time, in most cases, if you find the target object is far away from you, you can call moveahead, moveleft and move right multiple times.
2. If you find yourself seems to be stuck, you can lookdown to see if there's any object above or below you, you can also rotate to see if there's any object behind you.

Your response should first step-by-step reason about your state and plan, enclosed in <think>...</think>, followed by your action(s) in <answer>...</answer>.

You can take up to 5 action(s) at a time, separated by ','.
\end{promptbox}
Training ran for 150 GRPO steps with a sparse reward structure: $+10.0$ for stopping within 1.5m of the target, and $+0.5$ per turn for correct formatting. Collisions or illegal actions receive no explicit penalty.

Common RL hyperparameters are given in Table~\ref{tab:rl_hparams}.


\section{Generalization to Additional VLM Architectures}
\label{sec:appendix_arch}

\subsection{RL Dynamics on Qwen3-VL-4B}
\label{sec:appendix_qwen3_rl}

We train Qwen3-VL-4B with multi-turn RL in FrozenLake using the same setup and
VQA and probing configuration as for Qwen2.5-VL-3B.
Table~\ref{tab:qwen3_rl} reports VQA and probe accuracy
across 11 checkpoints. These additional RL results are limited to FrozenLake.

\begin{table}[H]
  \centering
  \small
  \setlength{\tabcolsep}{3pt}
  \renewcommand{\arraystretch}{1.12}
  \begin{tabular*}{\linewidth}{@{\extracolsep{\fill}}crrrr@{}}
    \toprule
    \textbf{Checkpoint} & \textbf{VQA} & \makecell{\textbf{Vision}\\\textbf{encoder}} & \makecell{\textbf{LLM}\\\textbf{L17}} & \makecell{\textbf{LLM}\\\textbf{L35}} \\
    \midrule
    0 & 42.22 & 72.22 & 78.89 & 88.89 \\
    1 & 44.56 & 72.78 & 79.44 & 89.44 \\
    2 & 46.11 & 73.89 & 80.56 & 90.00 \\
    3 & 45.89 & 71.67 & 80.00 & 89.44 \\
    4 & 47.44 & 72.22 & 81.11 & 90.00 \\
    5 & 48.00 & 74.44 & 81.67 & 91.11 \\
    6 & 49.11 & 71.67 & 82.78 & 90.56 \\
    7 & 49.89 & 72.78 & 83.89 & 91.67 \\
    8 & 50.11 & 70.56 & 82.22 & 91.11 \\
    9 & 51.00 & 71.67 & 84.44 & 91.67 \\
    10 & 51.89 & 71.11 & 85.00 & 91.67 \\
    \bottomrule
  \end{tabular*}
  \caption{VQA and probe accuracy (\%) of Qwen3-VL-4B across RL checkpoints in FrozenLake.}
  \label{tab:qwen3_rl}
\end{table}

\subsection{Attention Bias Across Architectures}

We test whether the plugin transfers to three additional backbones: Qwen3-VL-4B,
LLaVA-v1.6-Mistral-7B, and Gemma-3-4B.

\paragraph{Architecture-Specific Adaptation.}
The core challenge across architectures is localising image tokens in the input
sequence; the bias injection itself is identical once a per-token binary mask is
available.
Qwen3-VL brackets image tokens with explicit \texttt{vision\_start}/\texttt{vision\_end}
delimiters, so a linear scan over \texttt{input\_ids} suffices.
LLaVA-v1.6-Mistral uses a single \texttt{<image>} placeholder, with vision
features expanded internally; the true sequence length seen by each transformer
layer exceeds \texttt{input\_ids}. The plugin recovers this length via a one-shot
forward pre-hook and distributes the delta across placeholders to locate image tokens.
Gemma-3 expands each image into repeated \texttt{image\_token\_id} tokens at
tokenisation time, so image positions are found by a simple equality check against
\texttt{image\_token\_index}.
In all three cases, the binary mask is scaled by~$\alpha$, reshaped to
\texttt{(B,\,1,\,1,\,S)}, and added to the causal attention mask; during decoding
it is zero-padded for new tokens.

For each backbone the plugin is trained on the VQA training set with base model
parameters frozen, following the same procedure as the main paper; all other
settings are unchanged.

Table~\ref{tab:arch_intervention} reports VQA accuracy before (``O'') and after
(``P'') plugin insertion. The plugin improves accuracy across the evaluated
backbones, providing evidence for its applicability to additional VLM families.
These plugin evaluations are distinct from validation of RL-induced representational changes.

\end{document}